\documentclass{article}

\usepackage[numbers]{natbib}

\usepackage[final]{neurips_2023}
\usepackage[titletoc,toc,page]{appendix}
\usepackage{graphicx}
\usepackage{placeins}

\usepackage[utf8]{inputenc} % allow utf-8 input
\usepackage[T1]{fontenc}    % use 8-bit T1 fonts
\usepackage{hyperref}       % hyperlinks
\usepackage{url}            % simple URL typesetting
\usepackage{booktabs}       % professional-quality tables
\usepackage{amsfonts}       % blackboard math symbols
\usepackage{nicefrac}       % compact symbols for 1/2, etc.
\usepackage{microtype}      % microtypography
\usepackage{xcolor}         % colors
\usepackage{float}

\usepackage{fancyhdr}
\title{FloodBrain: Flood Disaster Reporting by Web-based Retrieval Augmented Generation with an LLM}
\author{%
  Grace Colverd%\thanks{%Use footnote for providing further information
    \\
    University of Cambridge \\ 
  \texttt{gb669@cam.ac.uk} \\
   \And
   Paul Darm%\footnotemark[1] %\thanks{%Use footnote for providing further information
    \\
  University of Strathclyde\\
  \texttt{paul.darm@strath.ac.uk} \\
  \AND
   Leonard Silverberg\\%\thanks{Use footnote for providing further information
  Trillium Technologies\\
  \texttt{leo@trillium.tech} \\
  \And
  Noah Kasmanoff\\%\thanks{Use footnote for providing further information
  New York University\\
  \texttt{nsk367@nyu.edu} \\
}

\begin{document}

\maketitle

\begin{abstract}
    Fast disaster impact reporting is crucial in planning humanitarian assistance. Large Language Models (LLMs) are well known for their ability to write coherent text and fulfill a variety of tasks relevant to impact reporting, such as question answering or text summarization. However, LLMs are constrained by the knowledge within their training data and are prone to generating inaccurate, or "hallucinated”, information. % We propose a specialized pipeline which uses information from websites to write web-based flood disaster impact reports about any event. This pipeline is implemented in our tool \textit{FloodBrain}(\href{https://floodbrain.com}{floodbrain.com}),\footnote{Demo and other information is available in the appendix.} specialized on providing information about flooding events. 
    To address this, we introduce a sophisticated pipeline embodied in our tool \textit{FloodBrain} (\href{https://floodbrain.com}{floodbrain.com}),\footnote{Demo and supplementary information can be found in the appendix.} specialized in generating flood disaster impact reports by extracting and curating information from the web.
    Our pipeline assimilates information from web search results to produce detailed and accurate reports on flood events. We test different LLMs as backbones in our tool and compare their generated reports to human-written reports on different metrics. Similar to other studies, we find a notable correlation between the scores assigned by GPT-4 and the scores given by human evaluators when comparing our generated reports to human-authored ones. Additionally, we conduct an ablation study to test our single pipeline components and their relevancy for the final reports. With our tool, we aim to advance the use of LLMs for disaster impact reporting and reduce the time for coordination of humanitarian efforts in the wake of flood disasters.

    %We present a comprehensive pipeline for extracting and curating information from web search results.  We also reproduce that scores assigned by GPT-4 on comparing our reports to human written reports has a high agreement with human provided scores. Additionally, we conduct an ablation study to test our single pipeline components and their relevancy for the final reports. 
    %Large Language Models (LLMs) are transforming the ways machine learning models are integrated with existing software and lowering the barrier to entry for such "AI-powered" applications. 
    % While many important caveats and issues are known and still under investigation, LLMs have enabled improved access and use for a variety of text-based problems, from content moderation to information retrieval, and more. 
    %In this work, we present \textit{FloodBrain} (\href{https://floodbrain.com}{floodbrain.com}),\footnote{Demo and other information is available in the appendix.} an LLM and accompanying software architecture specialized for humanitarian purposes. \textit{FloodBrain} is a tool that facilitates fast and accurate flood reporting for humanitarian organizations that are seeking information about major flooding events around the globe. Included in this work is a comprehensive long-form text generation pipeline, a chatbot which allows more flexible prompting, and optional integration with geospatial information when available. 
\end{abstract}

\section{Introduction}

%\begin{figure}
%    \centering
%    \includegraphics[width=10cm]{Latex/figures/FloodBrainui.png}
 %   \caption{Example report generated and presented on FloodBrain.com interactive %frontend.}
 %   \label{figure:FloodBrainui}
%\end{figure}

Floods are among the most devastating disasters around the globe. Between 2000 and 2019, over 1.65 billion people were affected by floods, comprising 41\% of total people affected by weather-related disasters  \cite{UNISDR_2020}. Global climate change contributes to the increased likelihood of floods due to the more extreme weather patterns it causes \cite{Milly_Wetherald_Dunne_Delworth_2002}. The last twenty years alone have seen the number of major floods more than double, from 1,389 to 3,254  \cite{UNISDR_2020}, with 24\% of the world's population exposed to floods \cite{Tellman_Sullivan_Kuhn_Kettner_Doyle_Brakenridge_Erickson_Slayback_2021}.  The exposure level to floods is especially high in low and middle-income countries, which are home to 89\% of the world's flood-exposed people \cite{A_2022}.  If human-induced climate change continues to escalate, the probability of encountering more severe weather events expands beyond the current high-risk areas. Therefore, the ability to respond to an increasing number of disasters will be critical.

% Consequently, we should anticipate more frequent instances of extreme flooding. This means that towns and cities previously experiencing floods as rare occurrences will now face the reality of them happening more frequently, no longer limited to 'once in a lifetime' events." [3]. 

\subsection{Large Language Models}

Large Language Models (LLMs) have had a significant impact on the machine learning and broader software development community in the past several months \cite{Eloundou_Manning_Mishkin_Rock_2023}. Consequently, a significant opportunity exists to explore how these tools can be used for building more effective systems and for automating rote tasks. 

One persistent problem is that LLMs can only know about events that were mentioned during their pretraining, which usually includes large corpora of text. Furthermore, LLMs can produce non-factual information by "hallucinating" facts in their output, as recent studies show that even state-of-the-art models can have patches in their internal knowledge representation \cite{berglund2023reversal}. Both of these are problematic in the context of disaster reporting. One method of mitigating these issues is to include verified text passages, that presumably contain the information to answer a specific question, in the model prompt together with the question. The reasoning is that with this technique the model can base its output on the provided context of the text passages and does not have to rely on its internal knowledge \cite{NEURIPS2020_Lewis}. 
%Due to the massive datasets upon which LLMs are trained and subsequently fine-tuned, the potential for prompting to produce valuable outputs can be extended to a variety of domains. 
%In this work, we aim to harness these possibilities for social good. By improving the curation, extraction, and dispatching of key insights regarding natural disaster data – in this case, floods – time is of the essence. We hope to better understand what opportunities exist for using LLMs and related approaches to create value in these sensitive situations.  
In this work, we use an LLM to summarize web-based content to assist with flood disaster reporting, with the aim of exploring how relevant information can be extracted, curated, and dispatched for humanitarian assistance. We name the resulting tool "FloodBrain".
% because of its focus on floods as well its automatic pipeline. 
We conduct an evaluation study to investigate how our generated reports compare to ground-truth human-generated reports with different LLMs as backbones for the pipeline and an ablation study to investigate the influence of different parts of our pipeline. 

\subsection{Disaster Reporting and Application Context}

When a disaster strikes, humanitarian agencies quickly mobilize to provide relief. While much good can be done with the initial funds these agencies have at their disposal, oftentimes an appeal to broader partners, such as the United Nations or local governments, is required to create sympathy for the situation and request for additional aid. A fundamental piece of these appeals is some sort of situational report that highlights the cause, impact, context, and future work needed for recovery. For brevity, we refer to these reports in the context of flooding disasters as "flood reports". Creating flood reports in a timely manner is essential for dispatching sufficient aid as quickly as possible. 

The creation of flood reports is a technical task that can be limited by several factors such as time available to the report maker, skill level, and access to the right data sources. Disaster relief agencies across the world use a varied set of report structures and conditions. Given the specialised nature of these flood reports, only a fraction of events are reported on, limited by the workforce available in these organisations. By automating the report-making process with a specialized pipeline we hope to expand the ability of organisations to respond to flooding events and decrease the time taken to distribute aid.  

% investigate how this could influence decision-making and faster disaster response time in the future. 

%In this work, we create a flood reporting solution that uses LLMs to ease the time and cognitive load required to write these reports. %We investigate the usefulness and trustworthiness of language models to this extent. Our breakthrough is a specialized pipeline for collaborative (human <> LLM) disaster report-making using state-of-the-art tools. 

During this research, we have collaborated and had discussions with people working in both the humanitarian aid process (World Food Programme, UNITAR's Operational Satellite Applications Programme) as well as in the flood mapping specific fields (RSS-Hydro). %We intend to keep developing these partnerships and relationships.

\section{Methodology}

%Floodbrain provides a user interface for report-making. The interface first allows a user to pick a known major "event" to generate a report on. These events are currently collected from three major disaster response agencies: the Copernicus Emergency Management Service (EMSR), the Global Disaster Alert and Coordination System (GDACS) and ReliefWeb, the humanitarian information portal from the United Nations Office for the Coordination of Human Affairs (OCHA). These services use their own criteria for assessing which floods are considered "major events" and reported on.  As a result, these events from separate sources provide good coverage of floods distributed globally and over time.  
% converting a given flooding disaster into one which is considered a major event. 

%Once FloodBrain is provided with an event code, three separate pipelines are activated and provide the user with a flood report within minutes, compared to the hours or days report generation would take if done without AI assistance.

%Since pre-trained LLMs have a temporal knowledge cut-off that does not include information pertaining to current events, simply "asking" will not produce satisfactory reports. Instead, FloodBrain uses trustworthy articles from the internet to augment its input and is then asked to perform simple tasks that lead to the desired results.

\subsection{Report Making}
For generating a flood report, FloodBrain follows a sequence of automated steps. The report generation is started by defining a key phrase, which includes the date and the location of a flooding disaster (e.g. "Flooding Paraguay Mar 2023"). This key phrase is then used to perform a web search to gather relevant websites from the internet. Textual data is extracted from each website, and stored as a "source" for the pipeline. 

\begin{figure}
    \centering
    \includegraphics[width=1\textwidth]{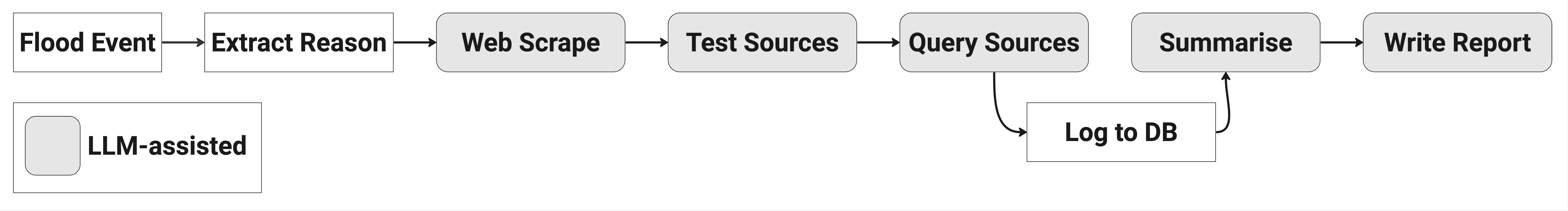}
    \caption{Sequential tasks in FloodBrain report generation.}
    \label{fig:report_maker_tasks}
\end{figure}

% For the web search, we use a client for Google search.
To gather more relevant sources we prompt an LLM to expand the original query into other relevant search queries. Each source is evaluated by an LLM to determine its relevancy to the original flood event and to filter out wrongly returned websites from the Google search. Relevant sources are then processed to extract information about the flood event. Each source is passed as context to an LLM tasked with answering a set of questions to encapsulate key themes required of a flood report, before being summarised into a final, coherent report. A flowchart of this process is illustrated in Figure \ref{fig:report_maker_tasks}.

The prompts used here were developed with feedback from domain experts from the World Food Programme and RSS-Hydro. Information on these prompts is available in Appendix \ref{tab:question_prompts}.

We provide a web-based user interface (UI) for accessing the FloodBrain report-making pipeline (floodbrain.com). The interface allows a user to pick a known major event to report on, which then defines the initial key phrase. These events are aggregated from three major disaster response agencies: the Copernicus Emergency Management Service (EMSR), the Global Disaster Alert and Coordination System (GDACS) and ReliefWeb, the humanitarian information portal from the United Nations Office for the Coordination of Human Affairs (OCHA). These services use their own criteria for converting a given flooding disaster into one which is considered a major event. As a result, these events from separate sources provide good coverage of floods distributed globally and over time. Full details of the UI and a demo of the report creation can be found in Appendix \ref{appendix:demo}. 

\subsection{Verification of Reports}

Trustworthiness is a crucial factor in flood reporting, and a transparent approach is needed if reports are to be used. Language models like LLMs are prone to hallucinations that diverge from factual accuracy, which hurts their reliability and applicability in real-world scenarios \cite{Kaddour_Harris_Mozes_Bradley_Raileanu_McHardy_2023}. To uphold the integrity of information used by practitioners, our pipeline incorporates a citation methodology. This feature enables users to "inspect" the data sources underpinning the generated report. For every source/question pairing, a text snippet from the source answering the respective question is logged within the %FloodBrain% 
tools database. Subsequently, these snippets are organized by question and displayed in the  UI, allowing users to review and validate the quality of the information encapsulated in the report. An example of this is given in Appendix \ref{appendix:Inspect_ui}.

\subsection{Evaluation}

We evaluate FloodBrain generated reports by comparing them to human-written reports from past disaster events. The human written reports are extracted from ReliefWeb, a humanitarian information portal, about disasters, which has flood reports for multiple larger events. An example ReliefWeb report is provided in Appendix \ref{appendix:rw_data}. This is meant to benchmark FloodBrain on past disasters, and see how applicable it is for future disasters, when no report is yet written. In this case, the FloodBrain report serves as a starting point for future report makers.

%We evaluate FloodBrain reports against existing flood reports using various techniques, utilizing ReliefWeb reports as ground truth data, while ensuring no ReliefWeb data is ingested into the FloodBrain pipeline for fairness. 
The evaluation encompasses three methods: ROUGE score comparison, LLM-assisted comparisons (G-EVAL), and human judgment. 

ROUGE scores (Recall-Orientated Understudy for Gisting Evaluation \cite{Lin_2004}), common for assessing machine-generated text quality by comparing it to a human-generated reference text \cite{Deutsch_Roth_2020}, offer an automated way to identify word similarity matching between generated and official reports, verifying region, impact, and event descriptions similarity. However ROUGE will not capture issues related to style, reports that capture more context, synonyms, or other hallucinations. We report on ROUGE recall so as to not penalise FloodBrain reports for containing more information (being longer) than ReliefWeb reports.  

% Recent research has shown promise in using LLMs to act as evaluation tools for grading generated text in addition to metrics like ROUGE. Liu et al. describe one approach of chain-of-thought prompting a LLM to asses generated text and name the methodology "G-EVAL" \cite{liu2023gpteval}. More specifically, an LLM is prompted to assess generated reports in comparison to source material for a specific metric (coherence, consistency, fluency, etc.) by returning a score between one and five, with five indicating the highest similarity. Notably, it was reported that the score assigned by the LLM had a significantly higher correlation with scores provided by human annotators than traditional metrics such as ROUGE. Similar, to measuring the recall score between the reference report and the generated report in this case the model is prompted to check if all facts mentioned in the reference are also included in the generated report. 

Recent studies indicate promise in LLMs as evaluation tools for grading generated text alongside metrics like ROUGE. Liu et al. introduced a methodology, "G-EVAL", using LLMs to assess generated text against source material on specific metrics like coherence, consistency, and fluency, scoring from one to five, with five being the highest similarity \cite{liu2023gpteval}. The LLM-assigned scores showed a notably higher correlation with human annotator scores than traditional metrics like ROUGE, effectively checking for comprehensive fact inclusion from the reference in the generated report. To validate G-EVAL in our situation, humans are subject to the same setup, and the corresponding results are compared.
\section{Results and discussion}

% \subsection{Language Models and Evaluation}

\textbf{G-EVAL and Human Evaluation}. Using a combination of G-EVAL, ROUGE and human evaluation, we compare the performance of different state-of-the-art LLMs in the FloodBrain pipeline. For these, experiments a dataset compromising 10 ReliefWeb reports together with 10 generated reports each for three different LLMs (GPT-4\footnote{\href{https://openai.com/research/gpt-4}{openai.com/research/gpt-4}}, GPT-3.5\footnote{\href{https://openai.com/blog/chatgpt}{openai.com/blog/chatgpt}}, and PaLM-Text-Bison \cite{anil2023palm}) is curated.
% Details on the prompt for the LLM call used in G-EVAL is given in the Appendix, found in Table \ref{}. 
For the G-EVAL evaluation, the original methodology was followed by using "GPT-4" to determine the score and prompting it ten times with a temperature setting set "1" to be able to approximate the token probability given the prior. To verify the findings of G-EVAL, a manual annotation round was conducted, involving four annotators assigned to perform an equivalent evaluation to G-EVAL checking for consistency between the reference and generated report. The outcomes of the evaluations and comparisons are presented in Table \ref{tab:comparison}.

% The prompt for the LLM call can be found in Appendix \paulcomment{placeholder}. Additionally, to verify the findings of G-EVAL, a manual annotation round was conducted, involving four annotators assigned to perform an equivalent evaluation to G-EVAL checking for consistency between the reference and generated report. For the G-EVAL evaluation, the original methodology was followed by using "GPT-4" to determine the score and prompting it ten times with a temperature setting set "1" to be able to approximate the token probability given the prior. For these, experiments a dataset compromising 10 ReliefWeb reports together with 10 generated reports each for three different LLMs (GPT-4\footnote{https://openai.com/research/gpt-4}, GPT-3.5\footnote{https://openai.com/blog/chatgpt}, and PaLM-Text-Bison \cite{palm2023}) of our pipeline was curated. The outcomes of the evaluations and comparisons are presented in Table \ref{tab:comparison}.

%\begin{table}[h!]
%    \centering
%    \caption{Consistency scores (out of 5) for various LLMs compared to ground truth ReliefWeb reports.}
%    \begin{tabular}{lccc}
%    \toprule

%    System & G-EVAL & Human  \\
%    \midrule
%    GPT-4  & 3.27 & 3.23 \\
%    GPT-3.5 & 2.96  & 2.78 \\
%    PaLM-Text-Bison  & 2.34 & 2.76  \\
%    \bottomrule
%    \end{tabular}

%    \label{tab:comparison}
%\end{table}

\begin{table}[h!]
    \centering
    \caption{G-EVAL scores (out of 5) and ROUGE Recall for various LLMs compared to ground truth ReliefWeb reports.}
    % Comparison of generated output of the system to ground-truth ReliefWeb reports. A higher score is better.}
    \begin{tabular}{lcccccc}
    \toprule
    & \multicolumn{2}{c}{G-EVAL} & \multicolumn{3}{c}{ROUGE Recall} \\
    \cmidrule(r){2-3} \cmidrule(r){4-6}
    System & Human & GPT-4 & 1 & 2 & L \\
    \midrule
    GPT-4 & 3.23 & 3.27 & 52.53 & 15.76 & 41.83 \\
    GPT-3.5 & 2.78 & 2.96 & 51.02 & 13.62 & 40.10 \\
    PaLM-Text-Bison & 2.76 & 2.34 & 41.43 & 10.20 & 32.08 \\
    \bottomrule
    \end{tabular}

    \label{tab:comparison}
\end{table}

GPT-4 exhibited superior performance compared to both GPT-3.5 (ChatGPT) and Google's PaLM-Text-Bison on average. Additionally, the G-EVAL scores presented a slightly lower score for the Google model and a slightly higher score for GPT-3.5 than the human scores. A similar trend can be observed in the ROUGE scores. Notably, the G-EVAL scores demonstrated the highest alignment with the assessments made by human annotators, which is illustrated in Table \ref{tab:correlation}, where the Pearson correlation metric is displayed. This higher correlation %G-EVAL significantly has a higher correlation with the human scores than the ROUGE scores, which
% is also in accordance with the results from the G-EVAL paper.
shows agreement with the seminal results from G-EVAL \cite{liu2023gpteval}.

% Given its performance on other benchmarks \paulcomment{source??}, the superior performance of GPT-4 is not surprising.

% although it could be interesting to investigate in which cases it performs better.
The application of G-EVAL demonstrates significant potential: its high correlation with human annotators suggests that G-EVAL or similar evaluation strategies could be instrumental in the evaluation of LLM generations and the refinement of the flood report generation pipeline. It could also aid in the selection of datasets for fine-tuning smaller models, to parallel the performance efficacy of GPT-4.  %In the future, a larger test set of human-generated reports vs. LLM-generated reports could be devised to further test the robustness of the methodology. 

%Through this metric we find that the OpenAI GPT models outperform PaLM as a base model for report making. with the largest of the three, GPT4, serving as the strongest performer for, human and G-EVAL, and recall based grading. Notably, the G-EVAL scores demonstrated the highest alignment with the assessments made by human annotators.This allignement is more distinctly illustrated in Table \ref{tab:correlation}, where the Pearson correlation metric is employed. This higher correlation %G-EVAL significantly has a higher correlation with the human scores than the ROUGE scores, which
%is in accordance with the results from the G-EVAL paper.

%Based on this result we expect this to serve as a possible benchmark for consideration for the use of potentially more freely available LLMs to be applied in this context for comparison. 

\begin{table}[h!]
 \caption{Pearson correlation scores of different evaluation metrics compared with human evaluation.}
    \centering
    \begin{tabular}{lcccc}
    \toprule
    Comparison & G-EVAL mean & ROUGE-1 & ROUGE-2 & ROUGE-L \\
    \midrule
    Human mean & 0.78 & 0.54 & 0.62 & 0.59 \\
    \bottomrule
    \end{tabular}
    \label{tab:correlation}
\end{table}

\textbf{Pipeline Ablation Study}. We conduct an ablation study to quantify the value of FloodBrain pipeline components, utilizing ROUGE recall for evaluation due to its lower API cost versus G-EVAL, and employing the Bison model as the LLM backbone. We also consider computational cost when evaluating the ablation. Removing the limitations of human annotators, the dataset is expanded to include 26 FloodBrain/ReliefWeb report pairs.

\begin{table}[H]
\caption{Recall results for ablation study on FloodBrain pipeline }
\label{tab:ablation_res}
\centering
\begin{tabular}{@{}llllll@{}}
\toprule
{Ablation Type}                                                       
                                                                                     & ROUGE-1 & ROUGE-2 & ROUGE-L & \\ \midrule
Full pipeline                                                                        & 36.64   & 8.38    & 29.46   &  \\
No enhanced search                                                                   & 34.35   & 7.86    & 27.34   &  \\
No source confirmation                                                               & 35.09   & 8.87    & 29.76   &   \\
\begin{tabular}[c]{@{}l@{}}No enhanced search \\ or source confirmation\end{tabular} & 33.85   & 8.22    & 27.49     \\ \bottomrule
\end{tabular}
\end{table}

The ablation study results are given in Table \ref{tab:ablation_res}. Removing LLM-assisted search decreases report quality across all ROUGE metrics: 6.3\% for ROUGE-1, 6.2\% for ROUGE-2, and 7.2\% for ROUGE-L.

Contrarily, removing the source relevancy check exhibits a mixed impact: ROUGE-1 declines by 4.2\%, while ROUGE-2 and ROUGE-L increase by 5.8\% and 1\% respectively. This could be attributed to two factors: a minor count of erroneously rejected sources including relevant information or closer phrase alignment to ReliefWeb reports, and the question/answer section acting as a secondary filter, filtering out irrelevant sources. The value of this pipeline step comes from accelerating report generation by eliminating extraneous context from downstream prompts, reducing computational cost by 59\%. In the complete pipeline, 41\% of sources pass the relevancy check (252 out of 611 for 26 reports), averting an additional 1,795 LLM API calls that would arise from the 359 failed sources continuing to the question/answer pipeline.

\section{Conclusion}
We developed a specialized pipeline for flood disaster reporting to showcase the potential application of LLMs in the field of humanitarian assistance.
Out of three tested LLMs, GPT-4 flood reports on average had the highest overlap with human-generated reports. Additionally in the evaluation, we confirm a high correlation between the scores
assigned by GPT-4 and the scores given by human evaluators when comparing our generated reports to human-authored ones. We found that augmenting our web search with LLM-generated additional queries further improves the agreement between our reports and human-generated reports while testing sources for relevance reduced computational costs by 59\%. Future studies could aim to further increase the quality of the flooding reports by working together with humanitarian relief agencies and analysing the content of the generated reports in more detail.

%FloodBrain offers a fast, accurate software pipeline for the generation of multi-modal reports on historic flood events. Our experiments have quantified the benefits of the full FloodBrain report-making pipeline, and proven the current SOTA performance of GPT-4 in humanitarian contexts. We have also validated that the G-EVAL methodology aligns well with human judgments concerning report quality in this context. 

% FloodBrain is currently limited to reporting on flooding events defined as disasters by external agencies. Given the lack of publicly shared methodology for what warrants a flood being deemed a major event and reported on, there is a risk of overlooking flooding events in areas of the world less attended to by these agencies. We hope to extend FloodBrain beyond these previously defined floods to improve access to flood reporting for these under-reported communities. Other example ways we hope this work can be expanded on are by exploring other humanitarian contexts or disaster types, and deploying a smaller and more efficient LLM specifically designed to this task, such as HUMBERT \cite{humbert}.

\textbf{Limitations and Future Work}. The FloodBrain UI currently reports only on externally recognized disaster-classified flooding events, risking oversight in less monitored regions due to unclear classification criteria of external agencies. Aiming to enhance flood reporting in under-reported communities, we aspire to broaden FloodBrain's scope beyond predefined floods, extend to other humanitarian or disaster contexts, and explore deploying a task-specific, efficient LLM.

% like HUMBERT \cite{humbert}.
% I think referencing new work in limitations seems pointless

\textbf{Ethical and Practical Considerations}. LLMs, while technologically advanced, pose environmental concerns due to their substantial computational, energy, and material demands, with GPT-3 training estimated to emit 502 tonnes CO2 \cite{luccioni2022estimating}. Additionally, employing generative AI from commercial "closed-source" entities risks ethical breaches as their datasets may not be ethically sourced, potentially generating malicious content or reinforcing existing biases.

% The environmental impact of operating LLMs is a notable concern, as their operation may contribute to climate change. LLMs epitomize the trade-off between technological advancement and environmental toll: the training of these large models demands substantial computational resources, which in turn necessitate significant energy and material inputs. The carbon emissions associated with training a model like GPT-3 are estimated to be 502 tonnes CO2 \cite{luccioni2022estimating}.

% When utilizing generative AI methodologies provided by commercial `closed-source' entities, there is no assurance that their training datasets are ethically sourced, posing a risk of generating malicious or harmful content or exacerbating existing biases embedded within their systems.

% The carbon cost of running LLMs is not certain but may contribute to climate change just by using this. Any progress in machine learning will come at a cost to the planet, and LLMs are no different, training large models requires a large amount of computational resources which in turn means energy and material. Carbon emissions of training a model like GPT-3 are estimated to be 502 tonnes \cite{luccioni2022estimating}. 
% When using generative AI methods from commercial `closed-source' companies, there is no guarantee that their training sets are based on ethically sourced data and that they will not generate malicious or harmful content, or amplify existing biases embedded in their systems.

Despite our efforts to curb hallucination risks via a human-in-the-loop \textit{inspect} methodology, skipping this step may lead to incorrect information in FloodBrain reports. Verification before official sharing is advised. This tool is designed for collaborative report writing between human and LLM, to be used cautiously.

% Though we have attempted to mitigate the risk of hallucination through the human-in-the-loop `inspect' methodology, if the user chooses not to do so, then incorrect information may be included in the FloodBrain report. Information should be verified before sharing in an official capacity. We built this workflow and tool to enable collaborative report writing between human<>LLM, and we intend for it to be used with caution.

\begin{ack}

This work has been enabled by \href{https://fdleurope.org}{Frontier Development Lab Europe} a public/private partnership between the European Space Agency (ESA), Trillium Technologies, the University of Oxford and leaders in commercial AI supported by Google Cloud and NVIDIA Corporation.  

FDL Europe and its outputs have been designed, managed and delivered by Trillium Technologies Ltd (trillium.tech).  Trillium is a research and development company with a focus on intelligent systems and collaborative communities for planetary stewardship, space exploration and human health. 

We express our gratitude to Google Cloud for providing extensive computational resources.

FDL Europe’s public/private partnership ensures that the latest tools and techniques in Artificial Intelligence (AI) and Machine Learning (ML) are applied to basic research priorities for planetary stewardship and disaster response, for all Humankind.
\end{ack}

% References follow the acknowledgments in the camera-ready paper. Use unnumbered first-level heading for
% the references. Any choice of citation style is acceptable as long as you are
% consistent. It is permissible to reduce the font size to \verb+small+ (9 point)
% when listing the references.
% Note that the Reference section does not count towards the page limit.
\medskip

\bibliographystyle{unsrtnat}
\bibliography{neurips_2023}

% \small

\appendix
\newpage
\section{FloodBrain UI }\label{appendix:UI}

FloodBrain provides a web-based UI for report-making. The interface first allows a user to pick a known major "event", to generate a report on. 

\subsection{FloodBrain Report Making Demo}\label{appendix:demo}
\begin{figure}
    \centering
    \includegraphics[width=1\linewidth]{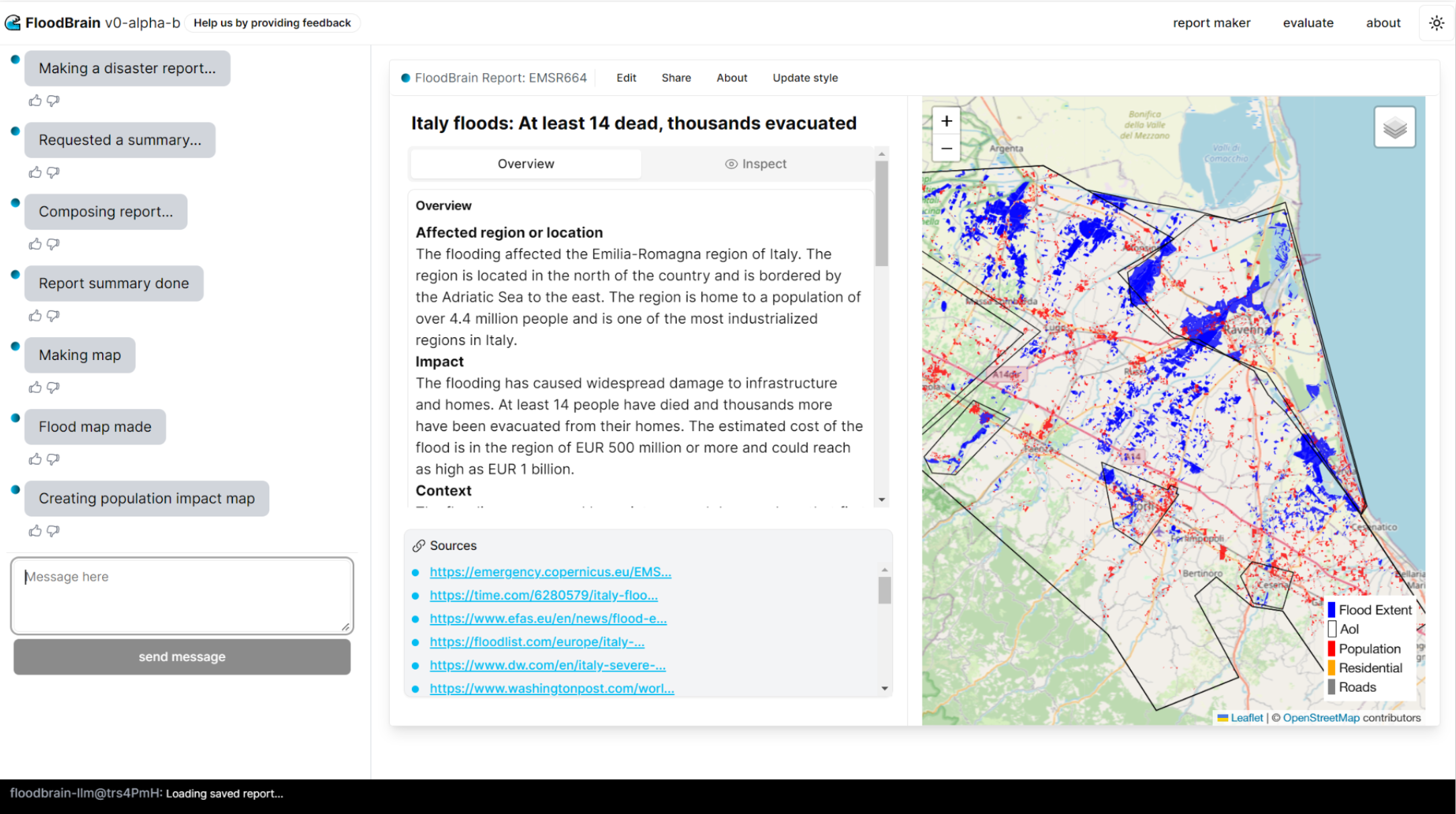}
    \caption{FloodBrain UI}
    \label{ui}
\end{figure}

Video link to report-making demo:
\href{https://www.youtube.com/watch?v=Dzgi9wn_0XU}{youtube.com/watch?v=Dzgi9wn\_0XU}
Figure \ref{ui} is a screen capture of the tool on the website: floodbrain.com.

\FloatBarrier

Once FloodBrain is provided with an event code, three separate pipelines are activated and provide the user with a flood report within minutes, compared to the hours or days report generation would take if done without AI assistance. 

The full UI includes a textual report, a flood extent map and a chatbot. 

\subsection{Textual Report}\label{apendix:qa_prompts}

The questions that form the backbone of the report-making are given in Table \ref{tab:question_prompts}. 

\begin{table}[h!]
\begin{tabular}{@{}l@{}}
\toprule
Questions                                                                                                                                \\ \midrule

 How many individuals were affected by the flooding, in which region or department, and on what date? \\
What are the regions affected by the flooding?                                                      \\
What specifically caused this flooding to occur?                                                      \\
 What is the timeline of this flooding?                                                        \\
 Are there any knock-on effects (disease, migration, etc.) reported as a result of the flood?         \\ \bottomrule
\end{tabular}
\label{tab:question_prompts}
\caption{Questions used in prompting for FloodBrain reports}
\end{table}

\subsection{Inspect UI}\label{appendix:Inspect_ui}

\begin{figure}[H]
    \centering
\includegraphics[width=.85\textwidth]{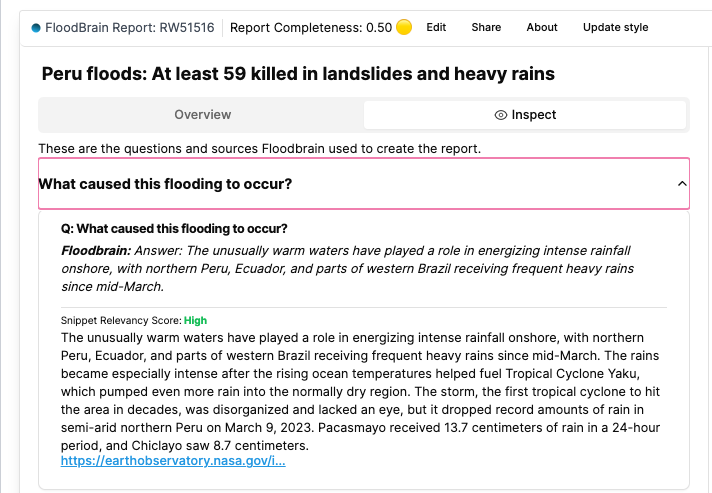}
    \caption{Inspect tab of the FloodBrain UI. Given a question, the user can see what the LLM responds with based on each source article.}
    \label{fig:inspect_tab}
\end{figure}

The Inspect tab allows the user of FloodBrain to review the data inputs into the report in order to validate and improve trustworthiness. An example of the Inspect UI is given in Figure \ref{fig:inspect_tab}. 

\subsection{Mapping}
FloodBrain augments its textual long-form reports with a geospatial report, aiming to enrich the contextual understanding of the flood report by visually representing the flooded area alongside other relevant geospatial data. We display flood extent,  affected population \cite{WorldPop}, affected points of interest \cite{osm} and affected infrastructure \cite{osm}.      

Several disaster response agencies offer mapping data in conjunction with their disaster charter activations. Whenever available, we leverage flood extent maps from these resources. In the absence of mapping data for certain flood events, we resort to employing existing flood segmentation algorithms to generate custom flood extent maps \cite{mateo-garcia_towards_2021}. Once the initial flood map is generated, we utilize the identified area of interest as a basis to gather mapping data from alternative sources such as OpenStreetMap. This additional data facilitates a more comprehensive comprehension of the disaster's impact, enabling a more nuanced analysis and response.

\subsection{Chatbot}

To enable further interrogation and interactivity alongside the Report and the Map, an LLM  Chatbot is also enabled.

The chatbot uses a ReAct \cite{Yao_Zhao_Yu_Du_Shafran_Narasimhan_Cao_2023} based reasoning system and is provided with the data sources collected in the reporting-making structure. Our users can then interact with the chatbot via a text interface and ask questions regarding the report.

\begin{figure}[H]
    \centering
    \includegraphics[width=1\linewidth]{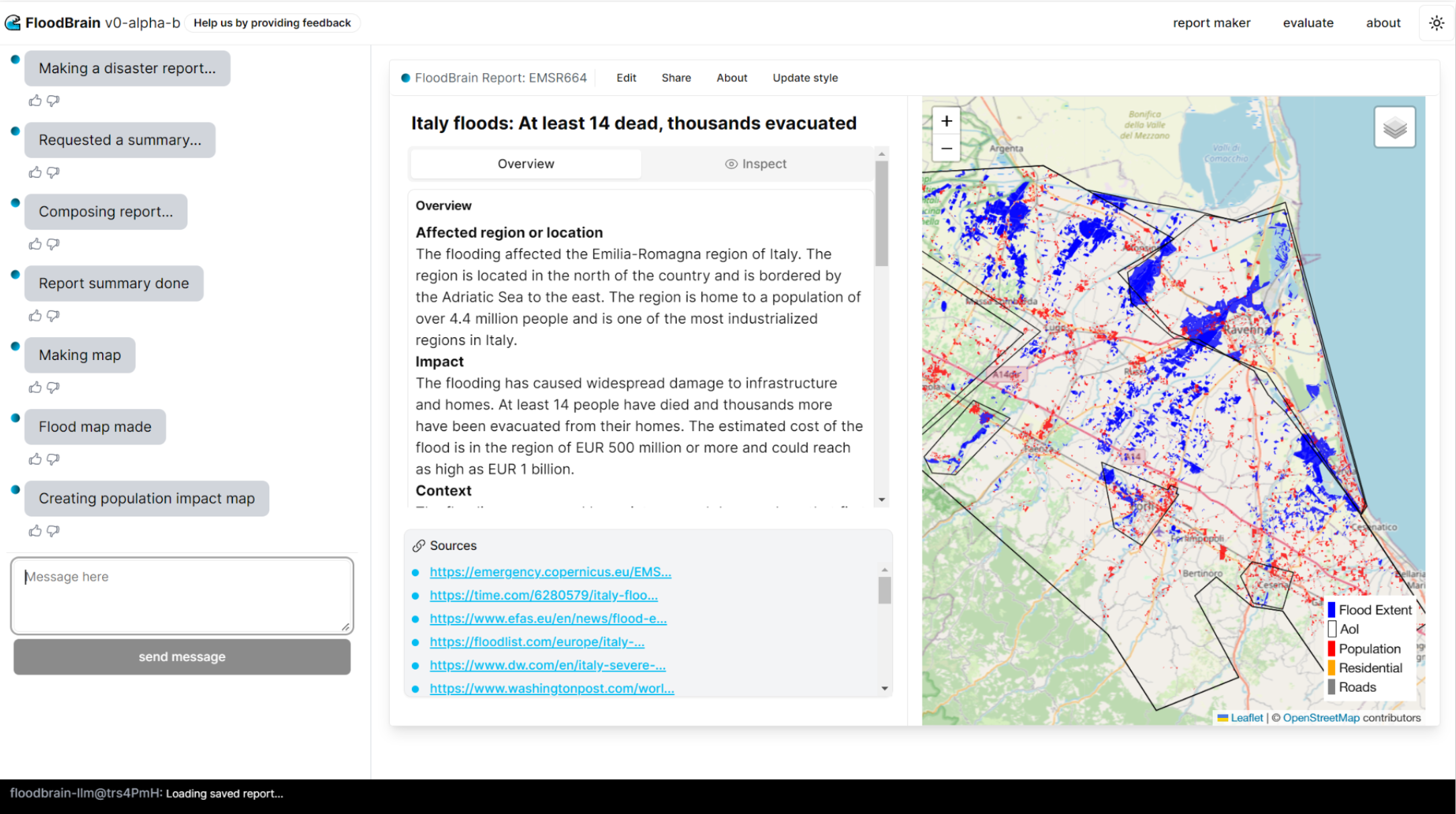}
    \caption{FloodBrain UI. In addition to the report created, a chatbot and flood extent map are also available.}
    \label{figure:ui}
\end{figure}

\subsection{Floodbrain Full Pipeline}
\begin{figure}[H]
    \centering
    \includegraphics[width=1\linewidth]{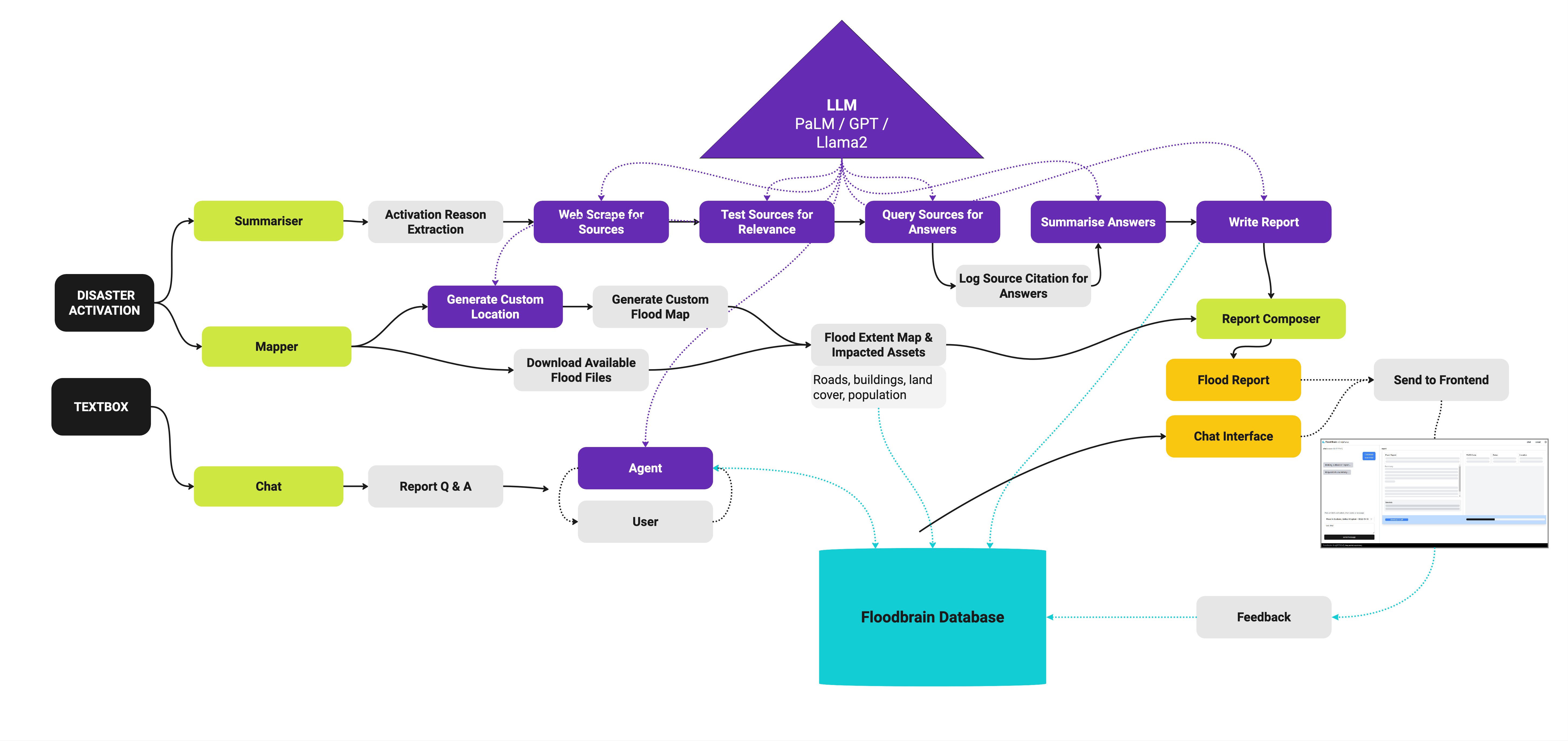}
    \caption{FloodBrain report-maker pipeline.}
    \label{pipeline}
\end{figure}

Figure \ref{pipeline} describes the full report-making pipeline, including mapping and chatbot. 

\section{ReliefWeb Reports}\label{appendix:rw_data}

\begin{figure}[H]
    \centering
    \includegraphics[width=0.8 \textwidth ]{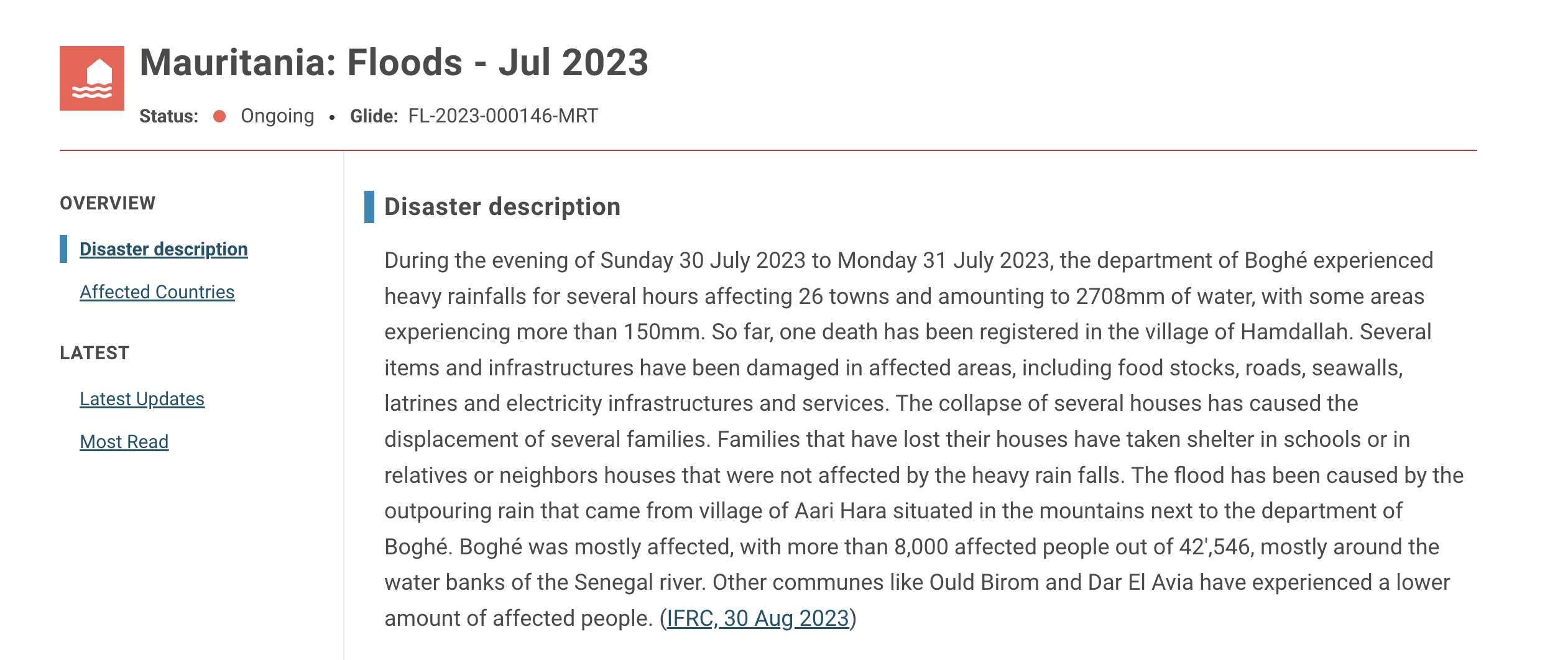}
    \caption{Example of ReliefWeb report for Mauritania}
    \label{fig:rw_example}
\end{figure}

In order to evaluate FloodBrain reports, we use ReliefWeb reports as ground truth data. ReliefWeb is a humanitarian information service provided by the United Nations Office for the Coordination of Humanitarian Affairs (OCHA). 
An example of such a report is given in Figure \ref{fig:rw_example} for the example of flooding in Mauritania in July 2023.

%%%%%%%%%%%%%%%%%%%%%%%%%%%%%%%%%%%%%%%%%%%%%%%%%%%%%%%%%%%%

\end{document}